\documentclass[11pt,a4paper]{article}
\usepackage[hyperref]{acl2018}
\usepackage{times}
\usepackage{latexsym}

\usepackage{url}
\usepackage{bm}

\usepackage{ifthen}
\usepackage{amsmath}
\usepackage{esvect}
\usepackage{graphicx}
\usepackage{tabularx}
\usepackage{soul}

\definecolor{mygray}{gray}{.9}
\definecolor{gray}{rgb}{0.5,0.5,0.5} 
\definecolor{green}{rgb}{0, 0.4, 0} 
\definecolor{orange}{rgb}{1, 0.5, 0}	
\definecolor{mahogany}{rgb}{0.75, 0.25, 0.0}
\definecolor{purple}{rgb}{0.6, 0, 0.6}
\definecolor{darkgreen}{rgb}{0, 0.4, 0.4} 
\definecolor{frenchblue}{rgb}{0.0, 0.45, 0.73}
\definecolor{myblue}{RGB}{74, 157, 255}
\definecolor{mygreen}{RGB}{0, 180, 100}

\newboolean{revising}
\setboolean{revising}{false}
\ifthenelse{\boolean{revising}}
{
	\newcommand{\ignore}[1]{}
    \newcommand{\Hsu}[1]{\textcolor{green}{#1}}

} {
	\newcommand{\ignore}[1]{}
    \newcommand{\Hsu}[1]{#1}

}

\aclfinalcopy 
\def\aclpaperid{335} 

\title{A Unified Model for Extractive and Abstractive Summarization \\using Inconsistency Loss}

\author{Wan-Ting Hsu$^1$, Chieh-Kai Lin$^1$, Ming-Ying Lee$^1$, Kerui Min$^2$, Jing Tang$^2$, Min Sun$^1$ \\
  $^1$ National Tsing Hua University, $^2$ Cheetah Mobile\\
  {\tt \{hsuwanting, axk51013, masonyl03\}@gapp.nthu.edu.tw,} \\
  {\tt \{minkerui, tangjing\}@cmcm.com, sunmin@ee.nthu.edu.tw}}

\date{}

\begin{document}
\maketitle

\begin{abstract}
We propose a unified model combining the strength of extractive and abstractive summarization. On the one hand, a simple extractive model can obtain sentence-level attention with high ROUGE scores but less readable. On the other hand, a more complicated abstractive model can obtain word-level dynamic attention to generate a more readable paragraph. In our model, sentence-level attention is used to modulate the word-level attention such that words in less attended sentences are less likely to be generated. Moreover, a novel inconsistency loss function is introduced to penalize the inconsistency between two levels of attentions. By end-to-end training our model with the inconsistency loss and original losses of extractive and abstractive models, we achieve state-of-the-art ROUGE scores while being the most informative and readable summarization on the CNN/Daily Mail dataset in a solid human evaluation. 
\end{abstract}

\begin{figure}[t!]
\begin{tabularx}{0.5\textwidth}{|X|}
\hline
  \scriptsize \textbf{Original Article:} \ul{McDonald's says}...... The company says \ul{it expects the new `Artisan Grilled Chicken' to be in its more than 14,300 U.S. stores by the end of next week, in products including a new sandwich, as well as existing sandwiches, wraps and salads. It says the biggest change is the removal of sodium phosphates}, which it said was used to keep the chicken moist, in favor of vegetable starch. \ul{The new recipe also does not use maltodextrin, which McDonald's said is generally used as a sugar to increase browning or as a carrier for seasoning.} Jessica Foust, director of culinary innovation at McDonald's, said the changes were made because customers said they want `simple, clean ingredients' they are familiar with...... And \textcolor{red}{\textit{Panera Bread has said}} it plans to purge artificial colors, flavors and preservatives from its food by 2016......\\
  \hline
  \scriptsize \textbf{Extractive Approach:} \textcolor{myblue}{\textbf{The company}} says it expects the new `Artisan Grilled Chicken' to be in its more than 14,300 U.S. stores by the end of next week, in products including a new sandwich, as well as existing sandwiches, wraps and salads. It says the biggest change is the removal of sodium phosphates, \textcolor{myblue}{\textbf{which it said was used to keep the chicken moist, in favor of vegetable starch}}. The new recipe also does not use maltodextrin, which McDonald's said is generally used as a sugar to increase browning or as a carrier for seasoning.\\
  \hline
  \scriptsize \textbf{Abstractive Approach:} McDonald's says it expects the new `Artisan Grilled Chicken' to be in its more than 14,300 U.S. stores by the end of next week. The company says the changes were made because customers said they want `simple, clean ingredients' they are familiar with. \textcolor{red}{\textit{McDonald's said}} it plans to purge artificial colors, flavors and preservatives from its food by 2016.\\
  \hline
  \scriptsize \textbf{Unified Approach:} \textcolor{mygreen}{\textbf{McDonald's says it expects}} the new `Artisan Grilled Chicken' to be in its more than 14,300 U.S. stores by the end of next week, in products including a new sandwich, as well as existing sandwiches, wraps and salads. It says the biggest change is the removal of sodium phosphates. The new recipe also does not use maltodextrin, which McDonald's said is generally used as a sugar to increase browning or as a carrier for seasoning.\\
  \hline
\end{tabularx}
\vspace{-4mm}
\caption{Comparison of extractive, abstractive, and our unified summaries on a news article. The extractive model picks most important but \textcolor{myblue}{\textbf{incoherent or not concise}} (see blue bold font) sentences. The abstractive summary is readable, concise but still \textcolor{red}{\textit{loses or mistakes some facts}} (see red italics font). The final summary rewritten from \underline{fragments} (see underline font) has the advantages from both extractive (importance) and abstractive advantage (\textcolor{mygreen}{\textbf{coherence}} (see green bold font)).}
\label{fig:ext_abs}
\end{figure}

\section{Introduction}
Text summarization is the task of automatically condensing a piece of text to a shorter version while maintaining the important points. The ability to condense text information can aid many applications such as creating news digests, presenting search results, and generating reports.
There are mainly two types of approaches: extractive and abstractive. Extractive approaches assemble summaries directly from the source text typically selecting one whole sentence at a time. In contrast, abstractive approaches can generate novel words and phrases not copied from the source text. \Hsu{Hence, abstractive summaries can be more coherent and concise than extractive summaries.}

Extractive approaches are typically simpler. They output the probability of each sentence to be selected into the summary. Many earlier works on summarization \cite{cheng2016neural,nallapati2016classify,nallapati2017summarunner,narayan2017neural,yasunaga2017graph} focus on extractive summarization. Among them, ~\citet{nallapati2017summarunner} have achieved high ROUGE scores.
On the other hand, abstractive approaches \cite{nallapati2016abstractive,see2017get,paulus2017deep,fan2017controllable,liu2017generative} typically involve sophisticated mechanism in order to paraphrase, generate unseen words in the source text, or even incorporate external knowledge. 
Neural networks~\cite{nallapati2017summarunner,see2017get} based on the attentional encoder-decoder model~\cite{bahdanau2014neural} were able to generate abstractive summaries with high ROUGE scores but suffer from inaccurately reproducing factual details and an inability to deal with out-of-vocabulary (OOV) words.
Recently, ~\citet{see2017get} propose a pointer-generator model which has the abilities to copy words from source text as well as generate unseen words. Despite recent progress in abstractive summarization, extractive approaches \cite{nallapati2017summarunner,yasunaga2017graph} and lead-3 baseline (i.e., selecting the first 3 sentences) still achieve strong performance in ROUGE scores. 

We propose to explicitly take advantage of the strength of state-of-the-art extractive and abstractive summarization and introduced the following unified model. 
Firstly, we treat the probability output of each sentence from the extractive model~\cite{nallapati2017summarunner} as sentence-level attention. Then, we modulate the word-level dynamic attention from the abstractive model~\cite{see2017get} with sentence-level attention such that words in less attended sentences are less likely to be generated.
In this way, extractive summarization mostly benefits abstractive summarization by mitigating spurious word-level attention.
Secondly, we introduce a novel inconsistency loss function to encourage the consistency between two levels of attentions. The loss function can be computed without additional human annotation and has shown to ensure our unified model to be mutually beneficial to both extractive and abstractive summarization. On CNN/Daily Mail dataset, our unified model achieves state-of-the-art ROUGE scores and outperforms a strong extractive baseline (i.e., lead-3). Finally, to ensure the quality of our unified model, we conduct a solid human evaluation and confirm that our method significantly outperforms recent state-of-the-art methods in informativity and readability.

To summarize, our contributions are \Hsu{twofold}:
\begin{itemize}
\item We propose a unified model combining sentence-level and word-level attentions to take advantage of both extractive and abstractive summarization approaches.
\item We propose a novel inconsistency loss function to ensure our unified model to be mutually beneficial to both extractive and abstractive summarization. The unified model with inconsistency loss \Hsu{achieves the best ROUGE scores on CNN/Daily Mail dataset and outperforms recent state-of-the-art methods in informativity and readability on human evaluation.}
\end{itemize}
\section{Related Work}

Text summarization has been widely studied in recent years. We first introduce the related works of neural-network-based extractive and abstractive summarization. Finally, we introduce a few related works with hierarchical attention mechanism.

\noindent\textbf{Extractive summarization.}
\citet{kaageback2014extractive} and \citet{yin2015optimizing} use neural networks to map sentences into vectors and select sentences based on those vectors. \citet{cheng2016neural}, \citet{nallapati2016classify} and \citet{nallapati2017summarunner} use recurrent neural networks to read the article and get the representations of the sentences and article to select sentences. \citet{narayan2017neural} utilize side information (i.e., image captions and titles) to help the sentence classifier choose sentences. \citet{yasunaga2017graph} combine recurrent neural networks with graph convolutional networks to compute the salience (or importance) of each sentence. While some extractive summarization methods obtain high ROUGE scores, they all suffer from low readability.

\begin{figure*} 	
\begin{center}
\includegraphics[width=\textwidth]{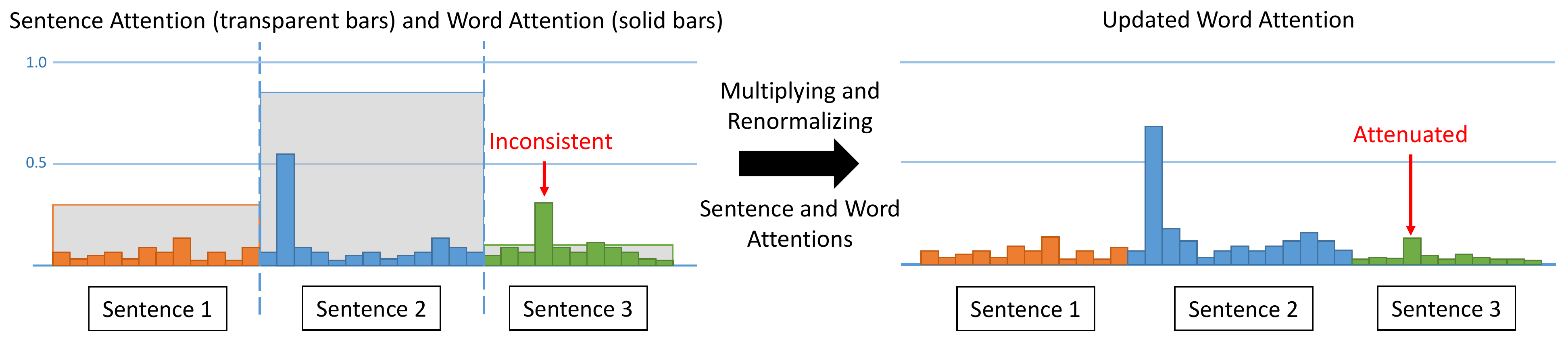}
\end{center}
\vspace{-5mm}
\caption{Our unified model combines the word-level and sentence-level attentions. Inconsistency occurs when word attention is high but sentence attention is low (see red arrow).}
\label{fig:combine}
\end{figure*}

\noindent\textbf{Abstractive summarization.} \citet{rush2015neural} first bring up the abstractive summarization task and use attention-based encoder to read the input text and generate the summary. Based on them, \citet{miao2016language} use a variational auto-encoder and \citet{nallapati2016abstractive} use a more powerful sequence-to-sequence model. 
Besides, \citet{nallapati2016abstractive} create a new article-level summarization dataset called CNN/Daily Mail by adapting DeepMind question-answering dataset \cite{hermann2015teaching}. \citet{ranzato2015sequence} change the traditional training method to directly optimize evaluation metrics (e.g., BLEU and ROUGE). \citet{gu2016incorporating}, \citet{see2017get} and \citet{paulus2017deep} combine pointer networks \cite{vinyals2015pointer} into their models to deal with out-of-vocabulary (OOV) words. \citet{chen2016distraction} and \citet{see2017get} restrain their models from attending to the same word to decrease repeated phrases in the generated summary. \citet{paulus2017deep} use policy gradient on summarization and state out the fact that high ROUGE scores might still lead to low human evaluation scores. \citet{fan2017controllable} apply convolutional sequence-to-sequence model 
 and design several new tasks for summarization. \citet{liu2017generative} achieve high readability score on human evaluation using generative adversarial networks. 

\noindent\textbf{Hierarchical attention.} Attention mechanism was first proposed by \citet{bahdanau2014neural}. \citet{yang2016hierarchical} proposed a hierarchical attention mechanism for document classification. 
\Hsu{We adopt the method of combining sentence-level and word-level attention in \citet{nallapati2016abstractive}. However, their sentence attention is dynamic, which means it will be different for each generated word. Whereas our sentence attention is fixed for all generated words. Inspired by the high performance of extractive summarization, we propose to use fixed sentence attention.}

Our model combines state-of-the-art extractive model \cite{nallapati2017summarunner} and abstractive model \cite{see2017get} by combining sentence-level attention from the former and word-level attention from the latter. Furthermore, we design an inconsistency loss to enhance the cooperation between the extractive and abstractive models.

\section{Our Unified Model}

We propose a unified model to combine the strength of both state-of-the-art extractor~\cite{nallapati2017summarunner} and abstracter~\cite{see2017get}. Before going into details of our model, we first define the tasks of the extractor and abstracter.

\noindent\textbf{Problem definition.}
The input of both extractor and abstracter is a sequence of words $\mathbf{w}=\left[w_1,w_2,...,w_m,...\right]$, where $m$ is the word index. The sequence of words also forms a sequence of sentences $\mathbf{s}=\left[s_1,s_2,...,s_n,...\right]$, where $n$ is the sentence index.
The $m^{th}$ word is mapped into the $n(m)^{th}$ sentence, where $n(\cdot)$ is the mapping function. The output of the extractor is the sentence-level attention $\bm{\beta}=\left[\beta_1,\beta_2,...,\beta_n,...\right]$, where $\beta_n$ is the probability of the $n^{th}$ sentence been extracted into the summary.
On the other hand, our attention-based abstractor computes word-level attention $\bm{\alpha}^t=\left[\alpha_1^t,\alpha_2^t,...,\alpha_m^t,...\right]$ dynamically while generating the $t^{th}$ word in the summary. The output of the abstracter is the summary text $\textbf{y}=\left[y^1,y^2,...,y^t,...\right]$, where $y^t$ is $t^{th}$ word in the summary.

In the following, we introduce the mechanism to combine sentence-level and word-level attentions in Sec.~\ref{sec.CA}. Next, we define the novel inconsistency loss that ensures extractor and abstracter to be mutually beneficial in Sec.~\ref{sec.IL}. We also give the details of our extractor in Sec.~\ref{sec:highlighter} and our abstracter in Sec.~\ref{sec:summarizer}. Finally, our training procedure is described in Sec.~\ref{sec:training}.

\subsection{Combining Attentions}\label{sec.CA}
Pieces of evidence (e.g., \citet{vaswani2017attention}) show that attention mechanism is very important for NLP tasks. Hence, we propose to explicitly combine the sentence-level $\beta_n$ and word-level $\alpha^t_m$ attentions by simple scalar multiplication and renormalization. The updated word attention $\hat{\alpha}^t_m$ is
\vspace{-5mm}
\begin{flalign}
&\hat{\alpha}^{t}_m = \frac{\alpha^t_m \times \beta_{n(m)}}{\sum_{m} \alpha^t_m\times \beta_{n(m)}}. 
\label{eq:m_alpha}
\end{flalign} \\[-3mm]
The multiplication ensures that only when both word-level $\alpha^t_m$ and sentence-level $\beta_n$ attentions are high, the updated word attention $\hat{\alpha}^t_m$ can be high.
Since the sentence-level attention $\beta_n$ from the extractor already achieves high ROUGE scores, $\beta_n$ intuitively modulates the word-level attention $\alpha^t_m$ to mitigate spurious word-level attention such that words in less attended sentences are less likely to be generated (see Fig.~\ref{fig:combine}). As highlighted in Sec.~\ref{sec:summarizer}, the word-level attention $\hat{\alpha}^t_m$ significantly affects the decoding process of the abstracter. Hence, an updated word-level attention is our key to improve abstractive summarization.

\subsection{Inconsistency Loss}\label{sec.IL}
Instead of only leveraging the complementary nature between sentence-level and word-level attentions, we would like to encourage these two-levels of attentions to be mostly consistent to each other during training as an intrinsic learning target for free (i.e., without additional human annotation). Explicitly, we would like the sentence-level attention to be high when the word-level attention is high. Hence, we design the following inconsistency loss,
\vspace{-3mm}
\begin{eqnarray}
L_{inc} = -\frac{1}{T}\sum_{t=1}^{T}\log (\frac{1}{|\mathcal{K}|}\sum_{m\in \mathcal{K}} \alpha^t_m \times \beta_{n(m)}),
\label{eq:loss_inc}
\end{eqnarray} \\[-4mm]
where $\mathcal{K}$ is the set of top K attended words and $T$ is the number of words in the summary.
This implicitly encourages the distribution of the word-level attentions to be sharp and sentence-level attention to be high. To avoid the degenerated solution for the distribution of word attention to be one-hot and sentence attention to be high, we include the original loss functions for training the extractor ( $L_{ext}$ in Sec.~\ref{sec:highlighter}) and abstracter ($L_{abs}$ and $L_{cov}$ in Sec.~\ref{sec:summarizer}).
Note that Eq.~\ref{eq:m_alpha} is the only part that the extractor is interacting with the abstracter. Our proposed inconsistency loss facilitates our end-to-end trained unified model to be mutually beneficial to both the extractor and abstracter.

\begin{figure} 	
\begin{center}
\includegraphics[width=0.47\textwidth]{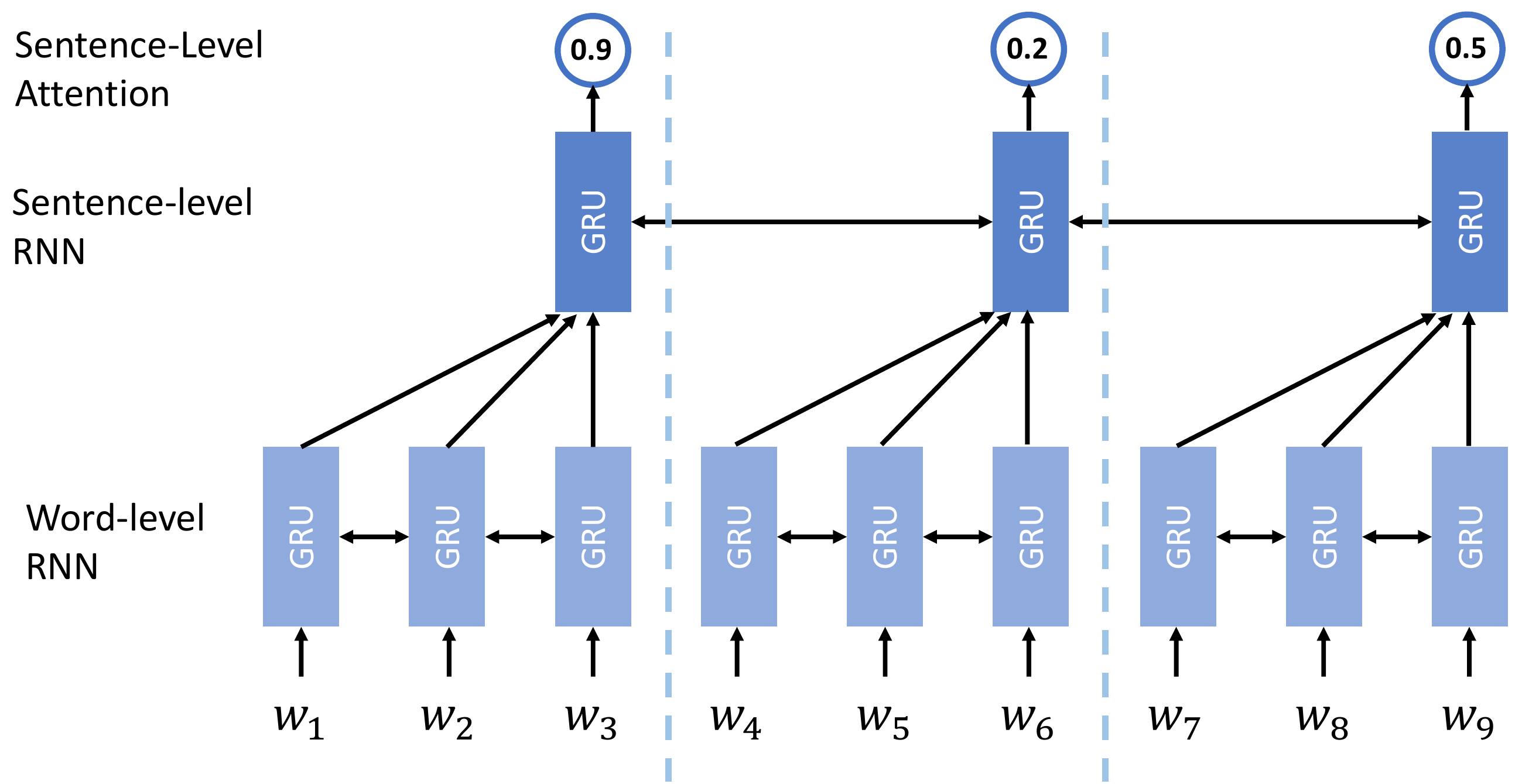}
\end{center}
\vspace{-6mm}
\caption{Architecture of the extractor. We treat the sigmoid output of each sentence as sentence-level attention $\in\left[0,1\right]$.}
\label{fig:ext_model}
\vspace{-2mm}
\end{figure}

\subsection{Extractor}
\label{sec:highlighter}

Our extractor is inspired by \citet{nallapati2017summarunner}. The main difference is that our extractor does not need to obtain the final summary. It mainly needs to obtain a short list of important sentences with a high recall to further facilitate the abstractor. We first introduce the network architecture and the loss function. Finally, we define our ground truth important sentences to encourage high recall.

\noindent\textbf{Architecture.} The model consists of a hierarchical bidirectional GRU which extracts sentence representations and a classification layer for predicting the sentence-level attention $\beta_n$ for each sentence (see Fig.~\ref{fig:ext_model}).

\noindent\textbf{Extractor loss.}
The following sigmoid cross entropy loss is used,
\vspace{-1mm}
\begin{flalign}
L_{ext}=-\frac{1}{N}\sum_{n=1}^{N} (g_n\log \beta_n+(1-g_n)\log(1-\beta_n)), 
\label{eq:loss_ext}
\vspace{-10mm}
\end{flalign} \\[-5mm]
where $g_n\in\{0,1\}$ is the ground-truth label for the $n^{th}$ sentence and $N$ is the number of sentences. When $g_n=1$, it indicates that the $n^{th}$ sentence should be attended to facilitate abstractive summarization.

\noindent\textbf{Ground-truth label.}
The goal of our extractor is to extract sentences with high informativity, which means the extracted sentences should contain information that is needed to generate an abstractive summary as much as possible. To obtain the ground-truth labels $\mathbf{g}=\{g_n\}_n$, first, we measure the informativity of each sentence $s_n$ in the article by computing the ROUGE-L recall score \cite{lin2004rouge} between the sentence $s_n$ and the reference abstractive summary $\mathbf{\hat{y}}=\{\hat{y}^t\}_t$. Second, we sort the sentences by their informativity and select the sentence in the order of high to low informativity. We add one sentence at a time if the new sentence can increase the informativity of all the selected sentences. Finally, we obtain the ground-truth labels $\mathbf{g}$ and train our extractor by minimizing Eq.~\ref{eq:loss_ext}. Note that our method is different from \citet{nallapati2017summarunner} who aim to extract a final summary for an article so they use ROUGE F-1 score to select ground-truth sentences; while we focus on high informativity, hence, we use ROUGE recall score to obtain as much information as possible with respect to the reference summary $\hat{\mathbf{y}}$.

\subsection{Abstracter}
\label{sec:summarizer}

\begin{figure} 	
\begin{center}
\includegraphics[width=0.48\textwidth]{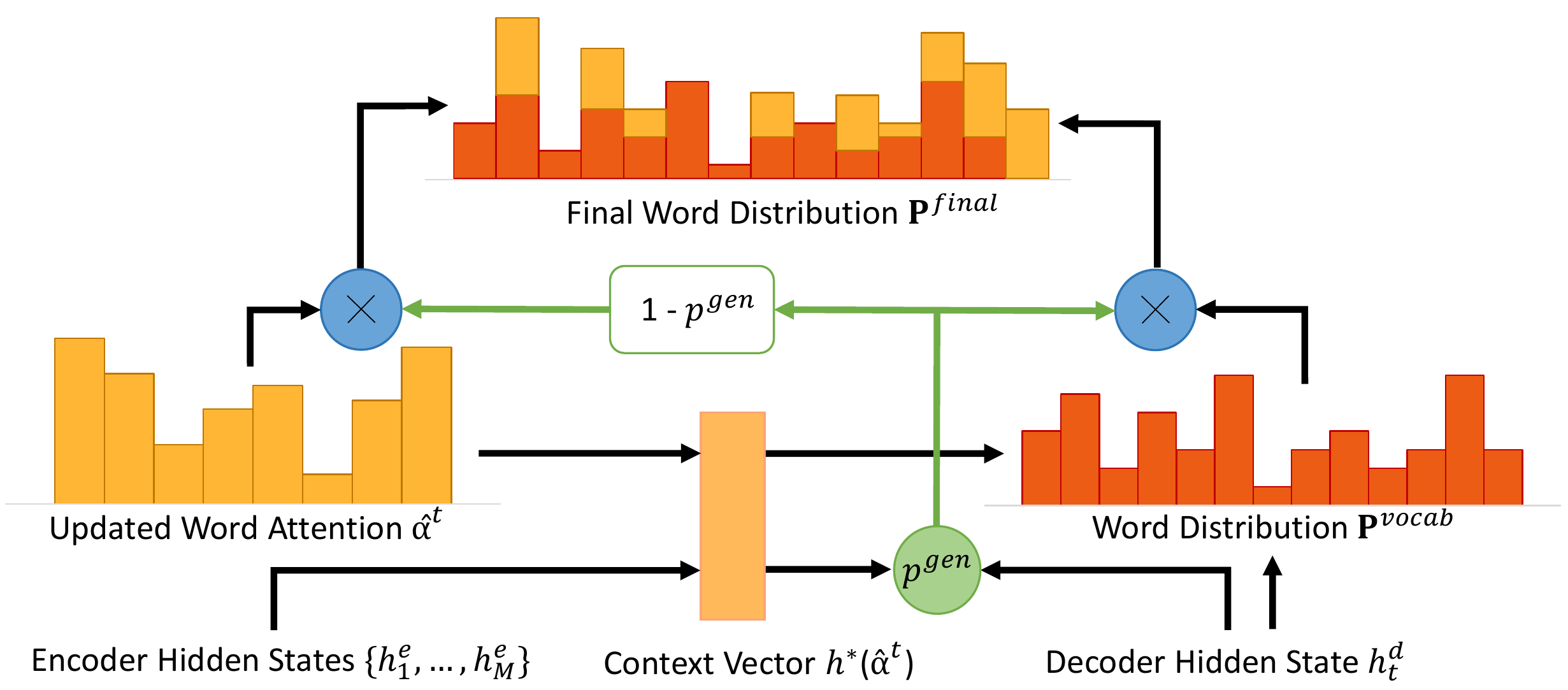}
\end{center}
\vspace{-4mm}
\caption{Decoding mechanism in the abstracter. In the decoder step $t$, our updated word attention $\bm{\hat{\alpha}}^t$ is used to generate context vector $h^*(\bm{\hat{\alpha}}^t)$. Hence, it updates the final word distribution $\mathbf{P}^{final}$.}
\vspace{-1mm}
\label{fig:abs_model}
\end{figure}

The second part of our model is an abstracter that reads the article; then, generate a summary word-by-word. We use the pointer-generator network proposed by \citet{see2017get} and combine it with the extractor by combining sentence-level and word-level attentions (Sec.~\ref{sec.CA}).

\noindent\textbf{Pointer-generator network.} The pointer-generator network \cite{see2017get} is a specially designed sequence-to-sequence attentional model that can generate the summary by copying words in the article or generating words from a fixed vocabulary at the same time. The model contains a bidirectional LSTM which serves as an encoder to encode the input words $\mathbf{w}$ and a unidirectional LSTM which serves as a decoder to generate the summary $\mathbf{y}$. 
For details of the network architecture, please refer to \citet{see2017get}. In the following, we describe how the updated word attention $\bm{\hat{\alpha}}^t$ affects the decoding process.

\noindent\textbf{Notations.}
We first define some notations. $h^e_m$ is the encoder hidden state for the $m^{th}$ word. $h^d_t$ is the decoder hidden state in step $t$. $h^*(\bm{\hat{\alpha}}^t)=\sum_{m}^{M}\hat{\alpha}_m^t\times h^e_m$ is the context vector which is a function of the updated word attention $\bm{\hat{\alpha}}^t$. $\mathbf{P}^{vocab}(h^*(\bm{\hat{\alpha}}^t))$ is the probability distribution over the fixed vocabulary before applying the copying mechanism.
\vspace{-2mm}
\begin{flalign}
\mathbf{P}&^{vocab}(h^*(\bm{\hat{\alpha}}^t))\\
&=\mathrm{softmax}(W_2(W_1[h^d_t,h^*(\bm{\hat{\alpha}}^t)]+b_1)+b_2),\nonumber
\end{flalign} \\[-6mm]
where $W_1$, $W_2$, $b_1$ and $b_2$ are learnable parameters. $\mathbf{P}^{vocab}=\{P^{vocab}_w\}_w$ where $P^{vocab}_w(h^*(\bm{\hat{\alpha}}^t))$ is the probability of word $w$ being decoded. $p^{gen}(h^*(\bm{\hat{\alpha}}^t))\in [0,1]$ is the generating probability (see Eq.8 in \citet{see2017get}) and $1-p^{gen}(h^*(\bm{\hat{\alpha}}^t))$ is the copying probability.

\noindent\textbf{Final word distribution.}
$P_w^{final}(\bm{\hat{\alpha}}^t)$ is the final probability of word $w$ being decoded (i.e., $y^t=w$). It is related to the updated word attention $\bm{\hat{\alpha}}^t$ as follows (see Fig.~\ref{fig:abs_model}),
\vspace{-1mm}
\begin{eqnarray}
P_w^{final}(\bm{\hat{\alpha}}^t)&=&p^{gen}(h^*(\bm{\hat{\alpha}}^t)) P^{vocab}_w(h^*(\bm{\hat{\alpha}}^t))\label{eq.abs}\\
&+&(1-p^{gen}(h^*(\bm{\hat{\alpha}}^t)))\sum_{m:w_m=w} \hat{\alpha}^t_m.\nonumber
\end{eqnarray} \\[-3mm]
Note that $\mathbf{P}^{final}=\{P_w^{final}\}_w$ is the probability distribution over the fixed vocabulary and out-of-vocabulary (OOV) words. Hence, OOV words can be decoded.
Most importantly, it is clear from Eq.~\ref{eq.abs} that $P_w^{final}(\bm{\hat{\alpha}}^t)$ is a function of the updated word attention $\bm{\hat{\alpha}}^t$.
Finally, we train the abstracter to minimize the negative log-likelihood:
\vspace{-3mm}
\begin{flalign}
L_{abs}=-\frac{1}{T}\sum_{t=1}^{T}\log{P_{\hat{y}^t}^{final}(\bm{\hat{\alpha}}^t)}~, 
\label{eq:loss_abst}
\end{flalign} \\[-4mm]
where $\hat{y}^t$ is the $t^{th}$ token in the reference abstractive summary.

\noindent\textbf{Coverage mechanism.} We also apply coverage mechanism \cite{see2017get} to prevent the abstracter from repeatedly attending to the same place. In each decoder step $t$, we calculate the coverage vector $\mathbf{c}^t=\sum_{t'=1}^{t-1} \bm{\hat{\alpha}}^{t'}$ which indicates so far how much attention has been paid to every input word. The coverage vector $\mathbf{c}^t$ will be used to calculate word attention $\bm{\hat{\alpha}}^t$ (see Eq.11 in \citet{see2017get}).
Moreover, coverage loss $L_{cov}$ is calculated to directly penalize the repetition in updated word attention $\bm{\hat{\alpha}}^t$:
\vspace{-2mm}
\begin{flalign}
L_{cov}=\frac{1}{T}\sum_{t=1}^T\sum_{m=1}^M\mathrm{min}(\hat{\alpha}^t_m,c^t_m)~. \label{eq:loss_cov} 
\end{flalign} \\[-4mm]
The objective function for training the abstracter with coverage mechanism is the weighted sum of negative log-likelihood and coverage loss.

\subsection{Training Procedure}
\label{sec:training}
We first pre-train the extractor by minimizing $L_{ext}$ in Eq.~\ref{eq:loss_ext} and the abstracter by minimizing $L_{abs}$ and $L_{cov}$ in Eq.~\ref{eq:loss_abst} and Eq.~\ref{eq:loss_cov}, respectively. When pre-training, the abstracter takes ground-truth extracted sentences (i.e., sentences with $g_n=1$) as input. To combine the extractor and abstracter, we proposed two training settings : (1) two-stages training and (2) end-to-end training.

\noindent\textbf{Two-stages training.}
In this setting, we view the sentence-level attention $\bm{\beta}$ from the pre-trained extractor as hard attention. The extractor becomes a classifier to select sentences with high attention (i.e., $\beta_n > \mathrm{threshold}$). We simply combine the extractor and abstracter by feeding the extracted sentences to the abstracter. Note that we finetune the abstracter since the input text becomes extractive summary which is obtained from the extractor.

\noindent\textbf{End-to-end training.}
For end-to-end training, the sentence-level attention $\bm{\beta}$ is soft attention and will be combined with the word-level attention $\bm{\alpha}^t$ as described in Sec.~\ref{sec.CA}. We end-to-end train the extractor and abstracter by minimizing four loss functions: $L_{ext}$, $L_{abs}$, $L_{cov}$, as well as $L_{inc}$ in Eq.~\ref{eq:loss_inc}. The final loss is as below:
\vspace{-2mm}
\begin{flalign}
\Hsu{L_{e2e}=\lambda_1 L_{ext}+ \lambda_2 L_{abs}+ \lambda_3 L_{cov}+ \lambda_4 L_{inc},}
\label{eq:loss_e2e}
\end{flalign} \\[-6mm]
\Hsu{where $\lambda_1$, $\lambda_2$, $\lambda_3$, $\lambda_4$ are hyper-parameters. In our experiment, we give $L_{ext}$ a bigger weight (e.g., $\lambda_1=5$) when end-to-end training with $L_{inc}$ since we found that $L_{inc}$ is relatively large such that the extractor tends to ignore $L_{ext}$.} 

\section{Experiments}

We introduce the dataset and implementation details of our method evaluated in our experiments.

\subsection{Dataset}

We evaluate our models on the CNN/Daily Mail dataset ~\cite{hermann2015teaching,nallapati2016abstractive,see2017get} which contains news stories in CNN and Daily Mail websites. Each article in this dataset is paired with one human-written multi-sentence summary. This dataset has two versions: \textit{anonymized} and \textit{non-anonymized}. The former contains the news stories with all the named entities replaced by special tokens (e.g., \verb|@entity2|); while the latter contains the raw text of each news story. We follow ~\citet{see2017get} and obtain the \textit{non-anonymized} version of this dataset which has 287,113 training pairs, 13,368 validation pairs and 11,490 test pairs.

\subsection{Implementation Details}

We train our extractor and abstracter with 128-dimension word embeddings and set the vocabulary size to 50k for both source and target text. We follow ~\citet{nallapati2017summarunner} and ~\citet{see2017get} and set the hidden dimension to 200 and 256 for the extractor and abstracter, respectively. We use Adagrad optimizer ~\cite{duchi2011adaptive} and apply early stopping based on the validation set. In the testing phase, we limit the length of the summary to 120.

\noindent\textbf{Pre-training.} We use learning rate 0.15 when pre-training the extractor and abstracter. For the extractor, we limit both the maximum number of sentences per article and the maximum number of tokens per sentence to 50 and train the model for 27k iterations with the batch size of 64. For the abstracter, it takes ground-truth extracted sentences (i.e., sentences with $g_n=1$) as input. We limit the length of the source text to 400 and the length of the summary to 100 and use the batch size of 16. We train the abstracter without coverage mechanism for 88k iterations and continue training for 1k iterations with coverage mechanism ($L_{abs}:L_{cov}=1:1$).

\noindent\textbf{Two-stages training.}
The abstracter takes extracted sentences with $\beta_n>$ 0.5, where $\bm{\beta}$ is obtained from the pre-trained extractor, as input during two-stages training. We finetune the abstracter for 10k iterations.

\noindent\textbf{End-to-end training.} During end-to-end training, we will minimize four loss functions (Eq.~\ref{eq:loss_e2e}) with \Hsu{$\lambda_1=5$ and $\lambda_2=\lambda_3=\lambda_4=1$}. We set K to 3 for computing $L_{inc}$. Due to the limitation of the memory, we reduce the batch size to 8 and thus use a smaller learning rate 0.01 for stability. The abstracter here reads the whole article. Hence, we increase the maximum length of source text to 600. \Hsu{We end-to-end train the model for 50k iterations.}

\begin{table*}[t!] 
\centering
\begin{tabular}{|l|c|c|c|}
\hline
  Method & ROUGE-1 & ROUGE-2 & ROUGE-L \\
  \hline
  pre-trained & 73.50 & 35.55 &  68.57 \\
  end2end w/o inconsistency loss & 72.97 & 35.11 & 67.99 \\
  end2end w/ inconsistency loss & \textbf{78.40} & \textbf{39.45} & \textbf{73.83} \\
    \hline \hline
  ground-truth labels & 89.23 & 49.36 & 85.46 \\
 \hline
\end{tabular}
\caption{ROUGE recall scores of the extracted sentences. \textit{pre-trained} indicates the extractor trained on the ground-truth labels. \textit{end2end} indicates the extractor after end-to-end training with the abstracter. Note that \textit{ground-truth labels} show the upper-bound performance since the reference summary to calculate ROUGE-recall is abstractive. \Hsu{All our ROUGE scores have a 95\% confidence interval with at most $\pm$0.33.}}
\label{table:highlight}
\end{table*}

\begin{table*}
\centering
\begin{tabular}{|l|c|c|c|}
\hline
  Method & ROUGE-1 & ROUGE-2 & ROUGE-L\\
  \hline
  \Hsu{HierAttn \cite{nallapati2016abstractive}$^{\ast}$} & \Hsu{32.75}& \Hsu{12.21} & \Hsu{29.01}\\
  \Hsu{DeepRL \cite{paulus2017deep}$^{\ast}$} & \Hsu{39.87}& \Hsu{15.82} & \Hsu{36.90}\\
  pointer-generator \cite{see2017get} & 39.53 & 17.28 & 36.38 \\
  GAN \cite{liu2017generative} & 39.92 & 17.65 & 36.71 \\
  \hline
  two-stage (ours) & 39.97 & 17.43 & 36.34 \\
  end2end w/o inconsistency loss (ours) & 40.19 & 17.67 & 36.68 \\
  end2end w/ inconsistency loss (ours)& \textbf{40.68} & \textbf{17.97} & \textbf{37.13}\\
  \hline
  lead-3 \cite{see2017get} & 40.34 & 17.70 & 36.57\\
  \hline
\end{tabular}
\caption{ROUGE F-1 scores of the generated abstractive summaries on the CNN/Daily Mail test set. Our two-stages model outperforms pointer-generator model on ROUGE-1 and ROUGE-2. In addition, our model trained end-to-end with inconsistency loss exceeds the lead-3 baseline. \Hsu{All our ROUGE scores have a 95\% confidence interval with at most $\pm$0.24. `$^{\ast}$' indicates the model is trained and evaluated on the anonymized dataset and thus is not strictly comparable with ours.}}
\label{table:highlightsum}
\end{table*}

\section{Results}

Our unified model not only generates an abstractive summary but also extracts the important sentences in an article. Our goal is that both of the two types of outputs can help people to read and understand an article faster. Hence, in this section, we evaluate the results of our extractor in Sec.~\ref{sec:results_highlight} and unified model in Sec.~\ref{sec:results_highlightsum}. Furthermore, in Sec.~\ref{sec:results_human}, we perform human evaluation and show that our model can provide a better abstractive summary than other baselines.

\subsection{Results of Extracted Sentences} 
\label{sec:results_highlight}

To evaluate whether our extractor obtains enough information for the abstracter, we use full-length ROUGE recall scores\footnote{\Hsu{All our ROUGE scores are reported by the official ROUGE script. We use the {\tt pyrouge} package. \\ \url{https://pypi.org/project/pyrouge/0.1.3/}}} between the extracted sentences and reference abstractive summary. High ROUGE recall scores can be obtained if the extracted sentences include more words or sequences overlapping with the reference abstractive summary. For each article, we select sentences with the sentence probabilities $\beta$ greater than $0.5$. We show the results of the ground-truth sentence labels (Sec.~\ref{sec:highlighter}) and our models on the test set of the CNN/Daily Mail dataset in Table ~\ref{table:highlight}. Note that the ground-truth extracted sentences can't get ROUGE recall scores of 100 because reference summary is abstractive and may contain some words and sequences that are not in the article.
Our extractor performs the best when end-to-end trained with inconsistency loss.

\begin{table*}[t!]
\centering
\begin{tabular}{|l|c|c|c|}
\hline
  Method & informativity & conciseness & readability\\
  \hline
  DeepRL~\cite{paulus2017deep} & 3.23 & 2.97 & 2.85 \\
  pointer-generator~\cite{see2017get} & 3.18 &3.36&3.47 \\
  GAN~\cite{liu2017generative} & 3.22& 3.52 & 3.51 \\
  Ours& \textbf{3.58} & 3.40 & \textbf{3.70}\\
  \hline
  reference & 3.43 & \textbf{3.61} & 3.62\\
  \hline
\end{tabular}
\caption{Comparing human evaluation results with state-of-the-art methods.}
\label{table:human}
\end{table*}

\begin{table}[t!]
\centering
\begin{tabular}{|l|c|}
\hline
  Method & avg. $R_{inc}$\\
  \hline
  w/o incon. loss & 0.198\\
  w/ incon. loss & 0.042 \\
  \hline
\end{tabular}
\caption{Inconsistency rate of our end-to-end trained model with and without inconsistency loss.}
\label{table:incon_ratio}
\end{table}

\subsection{Results of Abstractive Summarization}
\label{sec:results_highlightsum}

We use full-length ROUGE-1, ROUGE-2 and ROUGE-L F-1 scores to evaluate the generated summaries. We compare our models (two-stage and end-to-end) with state-of-the-art abstractive summarization models ~\cite{nallapati2016abstractive,paulus2017deep, see2017get,liu2017generative} and a strong lead-3 baseline which directly uses the first three article sentences as the summary. Due to the writing style of news articles, the most important information is often written at the beginning of an article which makes lead-3 a strong baseline. The results of ROUGE F-1 scores are shown in Table~\ref{table:highlightsum}. 
We prove that with help of the extractor, our unified model can outperform pointer-generator (the \Hsu{third} row in Table~\ref{table:highlightsum}) even with two-stages training (the \Hsu{fifth} row in Table~\ref{table:highlightsum}). After end-to-end training without inconsistency loss, our method already achieves better ROUGE scores by cooperating with each other. Moreover, our model end-to-end trained with inconsistency loss achieves state-of-the-art ROUGE scores and exceeds lead-3 baseline. 

In order to quantify the effect of inconsistency loss, we design a metric -- inconsistency rate $R_{inc}$ -- to measure the inconsistency for each generated summary. For each decoder step $t$, if the word with maximum attention  belongs to a sentence with low attention (i.e., $\beta_{n(\mathrm{argmax}(\bm{\alpha}^t))}<\mathrm{mean}(\bm{\beta})$), we define this step as an inconsistent step $t_{inc}$. The inconsistency rate $R_{inc}$ is then defined as the percentage of the inconsistent steps in the summary. 
\vspace{-4mm}
\begin{flalign}
R_{inc}=\frac{\mathrm{Count}(t_{inc})}{T},
\end{flalign} \\[-5mm]
where $T$ is the length of the summary. The average inconsistency rates on test set are shown in Table~\ref{table:incon_ratio}. Our inconsistency loss significantly decrease $R_{inc}$ from about $20\%$ to $4\%$. An example of inconsistency improvement is shown in Fig.~\ref{fig:ex_incon}.

\begin{figure}[t!]
\centering
\begin{tabularx}{0.48\textwidth}{|X|}
\hline
  \scriptsize \textbf{Without inconsistency loss:} \\ \vspace{-5mm} \scriptsize If that was a tornado, it was one monster of one. Luckily, so far it looks like no one was hurt. \ul{With tornadoes touching down near Dallas on Sunday, \textbf{Ryan Shepard snapped a photo of a black cloud formation reaching down to the ground.}} \textcolor{red}{\textbf{He said it was a tornado. It wouldn't be an exaggeration to say it looked half a mile wide.}} More like a mile, said Jamie Moore, head of emergency management in Johnson County, Texas. \ul{It could have been one the National Weather Service warned about in a tweet as severe thunderstorms drenched the area, causing street flooding.} (...)\\
  \hline
  \scriptsize \textbf{With inconsistency loss:} \\ \vspace{-5mm} \scriptsize If that was a tornado, it was one monster of one. Luckily, so far it looks like no one was hurt. \ul{With \textbf{tornadoes touching down near Dallas on Sunday, Ryan Shepard snapped a photo of a black cloud formation reaching down to the ground.}} He said it was a tornado.
It wouldn't be an exaggeration to say it looked half a mile wide.
More like a mile, said Jamie Moore, head of emergency management in Johnson County, Texas. \ul{\textbf{It could have been one the National Weather Service warned about in a tweet as} severe thunderstorms drenched the area, \textbf{causing street flooding.}} (...)\\
  \hline
\end{tabularx}
\caption{Visualizing the consistency between sentence and word attentions on the original article. We highlight word (bold font) and sentence (underline font) attentions. We compare our methods trained with and without inconsistency loss. Inconsistent fragments (see red bold font) occur when trained without the inconsistency loss.}
\label{fig:ex_incon}
\end{figure}

\begin{figure*}[t!] 
\centering
\begin{tabularx}{\textwidth}{|X|}
\hline
  \scriptsize \textbf{Original article (truncated):} \\ \vspace{-5mm} \scriptsize A chameleon balances carefully on a branch, waiting calmly for its prey... except that if you look closely, you will see that this picture is not all that it seems. For the `creature' poised to pounce is not a colourful species of lizard but something altogether more human. Featuring two carefully painted female models, it is a clever piece of sculpture designed to create an amazing illusion. It is the work of Italian artist Johannes Stoetter. Scroll down for video. Can you see us? Italian artist \textcolor{myblue}{\textbf{Johannes Stoetter has painted two naked women to look like a chameleon. The 37-year-old has previously transformed his models into frogs and parrots}} but this may be his most intricate and impressive piece to date. \textcolor{myblue}{\textbf{Stoetter daubed water-based body paint on the naked models to create the multicoloured effect, then intertwined them to form the shape of a chameleon.}} To complete the deception, the models rested on a bench painted to match their skin and held the green branch in the air beneath them. Stoetter can take weeks to plan one of his pieces and hours to paint it. Speaking about The Chameleon, he said: `I worked about four days to design the motif bigger and paint it with colours. The body painting took me about six hours with the help of an assistant. I covered the hair with natural clay to make the heads look bald.' Camouflage job: A few finishing touches are applied to the two naked models to complete the transformation. `There are different difficulties on different levels as in every work, but I think that my passion and love to my work is so big, that I figure out a way to deal with difficulties. My main inspirations are nature, my personal life-philosophy, every-day-life and people themselves.' However, the finished result existed only briefly before the models were able to get up and wash the paint off – with just a video and some photographs to record it. (...)\\
  \hline
  \scriptsize \textbf{Reference summary:} \\ \vspace{-5mm} \scriptsize Johannes Stoetter's artwork features two carefully painted female models. The 37-year-old has previously transformed models into frogs and parrots. Daubed water-based body paint on naked models to create the effect. Completing the deception, models rested on bench painted to match skin.\\
  \hline
  \scriptsize \textbf{DeepRL:} \\ \vspace{-5mm}  \scriptsize Italian artist Johannes Stoetter has painted female models to look like a chameleon. The 37-year-old has previously transformed his models into frogs and parrots but \textcolor{red}{\textit{this may be the work of Italian artist}}. He has painted nude models and it is a clever piece of sculpture designed to create an amazing illusion. \textcolor{red}{\textit{It is work of artist Johannes Stoetter.}}\\
  \hline
  \scriptsize \textbf{GAN:} \\ \vspace{-5mm} \scriptsize Italian artist Johannes Stoetter has painted two naked women to look like a chameleon. The 37-year-old has previously transformed his models into frogs and parrots \textcolor{red}{\textit{but this may be his most intricate and impressive piece to date}}.\\
  \hline
  \scriptsize \textbf{Pointer-generator:} \\ \vspace{-5mm} \scriptsize Italian artist Johannes Stoetter has painted two naked women to look like a chameleon. \textcolor{red}{\textit{It is the work of Italian artist Johannes Stoetter.}} Stoetter daubed water-based body paint on the naked models to create the multicoloured effect, then intertwined them to form the shape of a chameleon.\\
  \hline
  \scriptsize \textbf{Our unified model (with inconsistency loss):} \\ \vspace{-5mm} \scriptsize Johannes Stoetter has painted two naked women to look like a chameleon. The 37-year-old has previously transformed his models into frogs and parrots. Stoetter daubed water-based body paint on the naked models to create the multicoloured effect, then intertwined them to form the shape of a chameleon.\\
  \hline
\end{tabularx}
\vspace{-3mm}
\caption{Typical Comparison. Our model attended at the most important information (blue bold font) matching well with the reference summary; while other state-of-the-art methods generate repeated or less important information (red italic font).}
\label{fig:ex_human}
\end{figure*}

\subsection{Human Evaluation}
\label{sec:results_human}

We perform human evaluation on \Hsu{Amazon Mechanical Turk (MTurk)\footnote{\url{https://www.mturk.com/}}} to evaluate the informativity, conciseness and readability of the summaries. We compare our best model (end2end with inconsistency loss) with pointer-generator \cite{see2017get}, generative adversarial network \cite{liu2017generative} and deep reinforcement model \cite{paulus2017deep}. 
\Hsu{For these three models, we use the test set outputs provided by the authors}\footnote{\Hsu{\url{https://github.com/abisee/pointer-generator} and \url{https://likicode.com} for the first two. For DeepRL, we asked through email.}}. We randomly pick 100 examples in the test set. All generated summaries are re-capitalized and de-tokenized. Since \citet{paulus2017deep} trained their model on anonymized data, we also recover the anonymized entities and numbers of their outputs. 

\Hsu{We show the article and 6 summaries (reference summary, 4 generated summaries and a random summary) to each human evaluator. The random summary is a reference summary randomly picked from other articles and is used as a trap. We show the instructions of three different aspects as: (1) Informativity: how well does the summary capture the important parts of the article? (2) Conciseness: is the summary clear enough to explain everything without being redundant? (3) Readability: how well-written (fluent and grammatical) the summary is? The user interface of our human evaluation is shown in the supplementary material.} 

\Hsu{We ask the human evaluator to evaluate each summary by scoring the three aspects with 1 to 5 score (higher the better). We reject all the evaluations that score the informativity of the random summary as 3, 4 and 5. By using this trap mechanism, we can ensure a much better quality of our human evaluation. For each example, we first ask 5 human evaluators to evaluate. However, for those articles that are too long, which are always skipped by the evaluators, it is hard to collect 5 reliable evaluations. Hence, we collect at least 3 evaluations for every example. For each summary, we average the scores over different human evaluators.} 

The results are shown in Table~\ref{table:human}. \Hsu{The reference summaries get the best score on conciseness since the recent abstractive models tend to copy sentences from the input articles. However, our model learns well to select important information and form complete sentences so we even get slightly better scores on informativity and readability than the reference summaries.
}
We show a typical example of our model comparing with other state-of-the-art methods in Fig.~\ref{fig:ex_human}. \Hsu{More examples (5 using CNN/Daily Mail news articles and 3 using non-news articles as inputs) are provided in the supplementary material.}

\section{Conclusion}

We propose a unified model combining the strength of extractive and abstractive summarization. Most importantly, a novel inconsistency loss function is introduced to penalize the inconsistency between two levels of attentions. The inconsistency loss enables extractive and abstractive summarization to be mutually beneficial. By end-to-end training of our model, we achieve the best ROUGE-recall and ROUGE while being the most informative and readable summarization on the CNN/Daily Mail dataset in a solid human evaluation.

\Hsu{\section*{Acknowledgments}
We thank the support from Cheetah Mobile, National Taiwan University, and MOST 107-2634-F-007-007, 106-3114-E-007-004, 107-2633-E-002-001. We thank Yun-Zhu Song for assistance with useful survey and experiment on the task of abstractive summarization.}



\bibliographystyle{acl_natbib}

\end{document}